\documentclass[runningheads]{llncs}
\usepackage{float} 

\usepackage{amsmath, amsfonts, amssymb}
\usepackage{verbatim}
\usepackage{tabularx, booktabs}
\usepackage{graphicx}
\usepackage{tikz}
\usepackage{listings}
\usetikzlibrary{fit} 
 \usepackage{arydshln}
\newcommand{\squotes}[1]{`#1'}
\newcommand{\dquotes}[1]{``#1''}
\usepackage{graphicx}
\usepackage{subcaption}
\usepackage{makecell}

\usepackage{multirow}
\usepackage{ragged2e}  
\usepackage{booktabs}  
\usepackage{array, booktabs}
\usepackage{tabularx}
\usepackage{comment}
\usepackage[utf8]{inputenc} 
\usepackage[T1]{fontenc}    
\usepackage{hyperref}       
\usepackage{url}            
\usepackage{amsfonts}       
\usepackage{nicefrac}       
\usepackage{microtype}      
\usepackage{xcolor}         
\usepackage{amsmath}
\usepackage{colortbl}
\definecolor{Gray}{gray}{1}
\usepackage{xcolor}
\definecolor{shadecolor}{rgb}{0.9,0.9,0.9}

\usepackage[framemethod=TikZ]{mdframed}
\usepackage{lipsum}

\mdfdefinestyle{PromptStyle}{
  linecolor=blue,
  outerlinewidth=2pt,
  roundcorner=20pt,
  innertopmargin=\baselineskip,
  innerbottommargin=\baselineskip,
  innerrightmargin=20pt,
  innerleftmargin=20pt,
  backgroundcolor=gray!50!white}
\newcommand{\RandomForest}{\texttt{RF~}}
\newcommand{\LogisticRegression}{\texttt{LR~}}
\newcommand{\MLPClassifier}{\texttt{MLP~}}
\newcommand{\KNeighbors}{\texttt{KNN~}}
\newcommand{\XGBoost}{\texttt{XGB~}}
\newcommand{\AdaBoost}{\texttt{AdaBoost~}}

\definecolor{bluegreen}{rgb}{0,0,0} 

\definecolor{mycolor}{rgb}{0.5, 0.0, 0.5} 
\begin{document}
\title{Fairness of ChatGPT and the Role Of Explainable-Guided Prompts\thanks{Accepted to the workshop on \dquotes{\textit{Challenges and Opportunities of Large Language Models in Real-World Machine Learning Applications}}, COLLM@ECML-PKDD'23.}}
%
\titlerunning{Fairness of ChatGPT and the Role Of Explainable-Guided Prompts}
%
\author{Yashar Deldjoo} 
\authorrunning{Yashar Deldjoo}
\institute{Polytechnic University of Bari, Italy \\\email{deldjooy@acm.org}}

\maketitle              
\begin{abstract}
Our research investigates the potential of Large-scale Language Models (LLMs), specifically OpenAI's GPT, in credit risk assessment—a binary classification task. Our findings suggest that LLMs, when directed by judiciously designed prompts and supplemented with domain-specific knowledge, can parallel the performance of traditional Machine Learning (ML) models. Intriguingly, they achieve this with significantly less data—40 times less, utilizing merely 20 data points compared to the ML's 800. LLMs particularly excel in minimizing false positives and enhancing fairness, both being vital aspects of risk analysis. While our results did not surpass those of classical ML models, they underscore the potential of LLMs in analogous tasks, laying a groundwork for future explorations into harnessing the capabilities of LLMs in diverse ML tasks.
\end{abstract}
\section{Introduction and Context}
\noindent \textbf{Motivation.} Recent advancements in large language models such as OpenAI's GPT~\cite{brown2020language}, Google's PALM~\cite{chowdhery2022palm}, and Facebook's LaMDA~\cite{DBLP:journals/corr/abs-2201-08239} have redefined the landscape of Artificial Intelligence (AI). These behemoth models utilize billions of parameters and capitalize on the vastness of the internet data for training, leading to the generation of accurate and high-quality content. Large-scale Language Models (LLMs) have shown outstanding performance across tasks such as health diagnostics, job-seeking, and risk assessment, among others~\cite{chang2023survey,clavie2023large,qiu2023smile,alnuhait2023facechat,schaeffer2023emergent}. Given the transformative potential of these systems in decision-making across various contexts, their trustworthiness has drawn substantial attention. Unlike conventional ML models, LLMs leverage immense data scales, far surpassing those typically used for pretraining smaller or mid-scale models. This data, often sourced from the internet, mirrors societal norms but can also propagate prevalent societal biases. If unchecked, these biases can amplify, leading to biased outcomes that unfairly affect certain individuals or demographics.

A crucial aspect of harnessing these systems is through \textit{prompt engineering}, highlighted in this work. This technique mitigates the need for extensive dedicated training and offers system designers a measure of \squotes{control} over the model's behavior by enabling direct infusion of their insights into the learning process. While our focus is a case study on ChatGPT, the insights, and methodologies could potentially extend to other LLM types, an avenue we plan to explore in future work.

\vspace{1mm}
\begin{figure}
    \centering
    \includegraphics[trim=0 105pt 0 13pt, clip, width=0.90\linewidth]{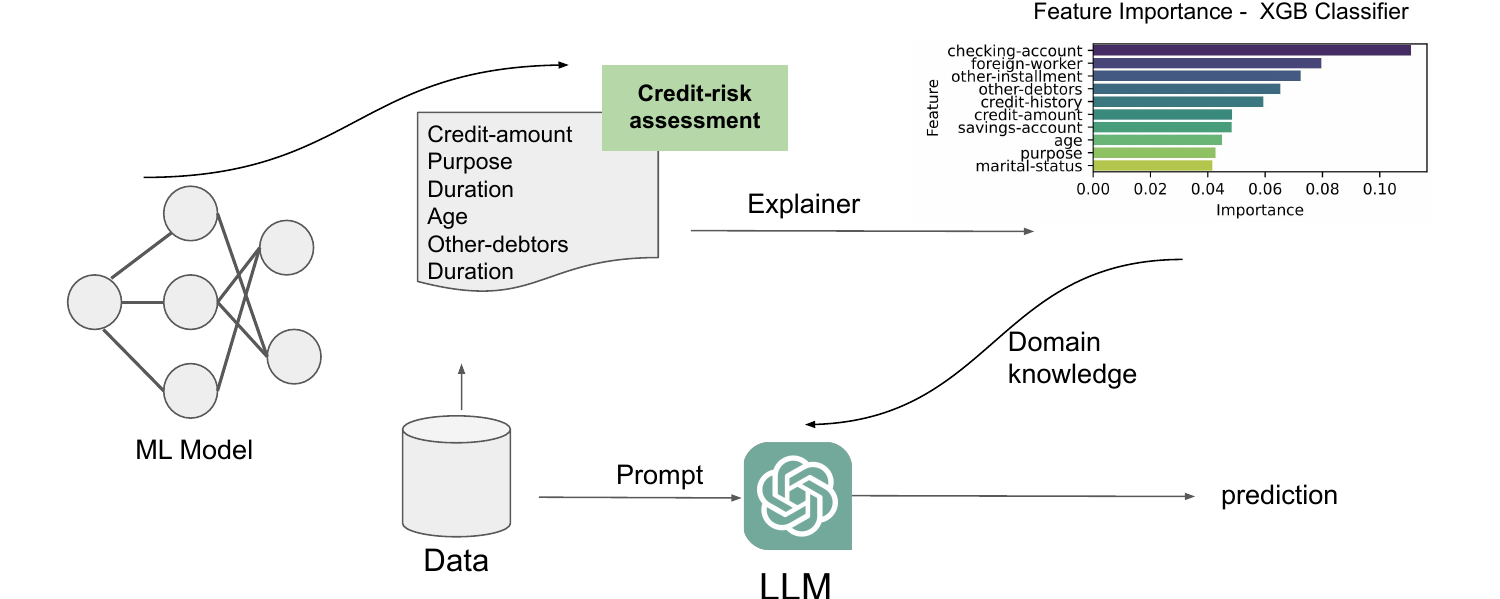}
\caption{Flowchart illustrating the conceptual framework of the paper}
\label{fig:flowchart}
\end{figure}

\noindent \textbf{Contributions.} This paper provides preliminary insights from an ongoing larger project aiming to address the challenges associated with the use of LLMs, particularly in decision-making processes through \textit{prompt engineering}. We demonstrate the ability to harness the potential of utilizing pre-trained models for downstream ML tasks, thereby eliminating the need for dedicated model training. By meticulously designing prompts that embody problem-specific instructions and contexts, we direct these models toward achieving desired objectives, such as enhancing prediction accuracy and minimizing risk (and/or mitigating unfair outcomes). Moreover, we underscore the significance of incorporating \textit{domain-specific knowledge} from experts, obtained through an apriori ML phase, as a powerful tool to improve the quality and effectiveness of the prediction tasks to a considerable degree. Our contributions can be summarized as follows:

\begin{itemize}
\item \textbf{OpenAI-ML Application:} We exemplify the application of OpenAI's GPT for specific ML tasks, focusing on credit risk assessment as a case study;
\item \textbf{Prompt Engineering:} We investigate the impact of different prompts and their parameters on the outcomes of ML tasks;
\item \textbf{Domain Knowledge Integration:} We propose a method for enhancing openAI-ML model accuracy by integrating optimal features, as identified by the ML models employed a priori. This demonstrates how leveraging feature importance  can boost model performance when used as domain knowledge;
\item \textbf{Bias of Classical-ML vs. OpenAI-ML:} Contrary to the approach of \cite{li2023fairness} that focuses on aggregate metrics, we scrutinize biases in OpenAI ML models, using gender as a case study. We assess gender fairness not through aggregate metrics but by comparing distributions via bootstrap sampling and evaluating results with statistical significance tests.

\end{itemize}

Our research is aimed at providing a guide for utilizing LLMs in ML tasks, with a primary focus on enhancing accuracy through prompt engineering and assessing its impact on fairness outcomes. We expand on previous work by Li et al.\cite{li2023fairness}, where fairness-based prompt engineering was conducted across several datasets. We have demonstrated how the accuracy of these systems can be substantially enhanced and supplemented this with a detailed fairness analysis using statistical measures. \footnote{The link to the system developed can be found in \url{https://github.com/yasdel/ChatGPT-FairXAI}.}

\vspace{-4mm}

\section{OpenAI-ML Framework for Credit Assessment Task}

We utilize ChatGPT-3.5-Turbo via the \textit{chat-completion} API, chosen for its outstanding text generation, speed, and cost-effectiveness. Converting the ML task into a \dquotes{chat conversation} with ChatGPT is pivotal for prediction, ensuring the model responds in binary format -- yes or no. Constructing prompts, which link the model and task, demands understanding of both context and capabilities. Effective prompts harness the model's comprehension skills for complex tasks, though their creation is challenging. Fig. \ref{fig:flowchart} visually depicts the process for designing prompts for downstream ML tasks.

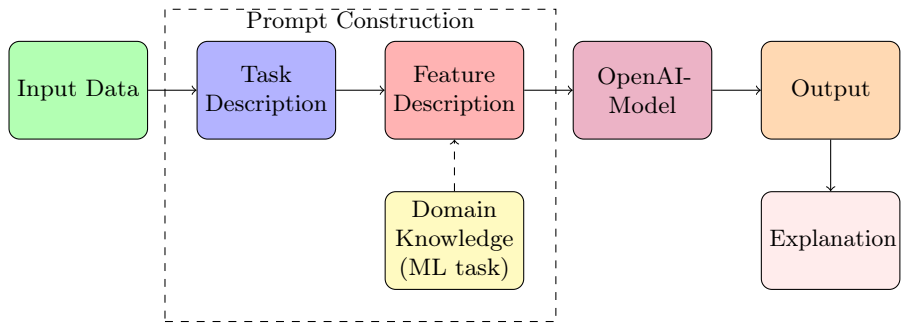
\begin{figure}[!h]
\centering
\begin{tikzpicture}[scale = 0.6, node distance = 1.8cm, auto, block/.style={rectangle, draw, text width=5em, text centered, rounded corners, minimum height=4em}]

    \node [block, fill=green!30] (input_data) {Input Data};
    \node [block, right of=input_data, node distance=2.5cm, fill=blue!30] (starting_description) {Task Description};
    \node [block, right of=starting_description, node distance=2.5cm, fill=red!30] (feature_description) {Feature Description};
    \node [block, below of=feature_description, node distance=2cm, fill=yellow!30] (domain_knowledge) {Domain Knowledge (ML task)};
    \node [block, right of=feature_description, node distance=2.5cm, fill=purple!30] (model) {OpenAI-Model};
    \node [block, right of=model, node distance=2.5cm, fill=orange!30] (output) {Output};
    \node [block, below of=output, node distance=2cm, fill=pink!30] (explanation) {Explanation};

    \begin{scope}[local bounding box=prompt_construction, dashed]
        \node [fit=(starting_description) (feature_description) (domain_knowledge), draw, dashed, inner sep=0.42cm, rectangle] {};
        \node [above of=prompt_construction, node distance=1.9cm] {Prompt Construction};
    \end{scope}

    \path [draw, ->] (input_data) -- (starting_description);
    \path [draw, ->] (starting_description) -- (feature_description);
    \path [draw, dashed, ->] (domain_knowledge) -- (feature_description);
    \path [draw, ->] (feature_description) -- (model);
    \path [draw, ->] (model) -- (output);
    \path [draw, ->] (output) -- (explanation);
\end{tikzpicture}
\caption{Diagram of the prompt creation and credit evaluation process using OpenAI. Note that we plan incorporate an explanation service via the API.}
\label{fig:flowchart}
\end{figure}

\vspace{-3mm}

\noindent \textbf{Prompt Construction.} This technique provides task-oriented instruction for the OpenAI model, delivering the necessary context. Our method starts with \dquotes{Part 1. Task Instruction}, where we guide the model on its task, then \dquotes{Part 2. In-context Examples} to boost predictions. In \dquotes{Part 3. Attribute Description}, we detail task-specific features. This is followed by \dquotes{Part 4. Integration of Domain Knowledge}, strategically incorporated to improve model comprehension and accuracy. The final stage is \dquotes{Part 5. Formulation of a Question/Problem}, framing the task or query at hand.  Note that the  \textit{Integration of Domain Knowledge} is strategically included to enhance the model's understanding and prediction accuracy (cf. Section~\ref{subsec:domain}).

\mdfdefinestyle{PartOneStyle}{
  linecolor=blue,
  outerlinewidth=2pt,
  roundcorner=20pt,
  innertopmargin=10pt,
  innerbottommargin=10pt,
  innerrightmargin=10pt,
  innerleftmargin=10pt,
  backgroundcolor=green!20!white,
  frametitle={\textbf{Part 1: Task Instruction}}
}
\mdfdefinestyle{PartTwoStyle}{
  linecolor=blue!60!white,
  outerlinewidth=2pt,
  roundcorner=20pt,
  innertopmargin=10pt,
  innerbottommargin=10pt,
  innerrightmargin=10pt,
  innerleftmargin=10pt,
  backgroundcolor=yellow!20!white,
  frametitle={\textbf{Part 2: In-Context Example}}
}
\mdfdefinestyle{PartThreeStyle}{
  linecolor=blue!60!white,
  outerlinewidth=2pt,
  roundcorner=20pt,
  innertopmargin=10pt,
  innerbottommargin=10pt,
  innerrightmargin=10pt,
  innerleftmargin=10pt,
  backgroundcolor=red!20!white,
  frametitle={\textbf{Part 3: Attribute Description}}
}
\mdfdefinestyle{PartFourStyle}{
  linecolor=blue!60!white,
  outerlinewidth=2pt,
  roundcorner=20pt,
  innertopmargin=10pt,
  innerbottommargin=10pt,
  innerrightmargin=10pt,
  innerleftmargin=10pt,
  backgroundcolor=orange!20!white,
  frametitle={\textbf{Part 4: Domain Knowledge Inetgration}}
}
\mdfdefinestyle{PartFiveStyle}{
  linecolor=blue,
  outerlinewidth=2pt,
  roundcorner=20pt,
  innertopmargin=10pt,
  innerbottommargin=10pt,
  innerrightmargin=10pt,
  innerleftmargin=10pt,
  backgroundcolor=purple!20!white,
  frametitle={\textbf{Part 5: Final Task Question}}
}

\begin{mdframed}[style=PartOneStyle]
Evaluate the credit risk based on given attributes. If good, respond with '1', if bad, respond with '0'.
\end{mdframed}

\begin{mdframed}[style=PartTwoStyle]
Here's an example: a customer with a good checking account history, a credit duration of 12 months, no history of bad credit, and a purpose of car loan, requested a credit amount of \$5000. The system evaluated the risk as good and responded with '1'.
\end{mdframed}

\begin{mdframed}[style=PartThreeStyle]
Consider each attribute:

\begin{itemize}
 \item Checking-account: Existing status
 \item Duration: Credit duration (months)
 \item Credit-history
 \item Purpose: (car, furniture, etc.)
 \item Credit-amount
\end{itemize}
\end{mdframed}

\begin{mdframed}[style=PartFourStyle]
Domain Knowledge: 

\textbf{dk2}: Important features in evaluating credit risk often include Checking-account, Foreign-worker, Other-installment, Other-debtors, Credit-history, Credit-amount, and Savings-account.

\textbf{dk3}: The order of features is important in evaluating credit risk. Important features in evaluating credit risk involve assessing each feature sequentially, starting with the Checking-account, then moving to Foreign-worker, Other-installment, Other-debtors, Credit-history, Credit-amount, Savings-account, Age, Purpose, and finally Duration.
\end{mdframed}

\begin{mdframed}[style=PartFiveStyle]
Based on the provided inputs and domain knowledge, is the credit risk good (\squotes{1} or bad \squotes{0})?
\end{mdframed}

\subsection{Domain knowledge Integration}
\label{subsec:domain}

In the context of ML tasks, domain knowledge is typically provided by the domain expert, such as a bank expert in the case of credit risk assessment. However, this domain knowledge can also be simulated by an ML model, which learns not only the relevance of individual features but also their interconnections. To evaluate the impact of this domain knowledge on task performance, a wide range of ML models were utilized as part of the domain knowledge for the OpenML prediction task. Particularly, we introduced three categories of domain knowledge, as detailed in Table \ref{tab:my_label}: \squotes{dk0} (\texttt{prompt-0}) represents a base case with no domain knowledge and learning is purely data-driven, \squotes{Odd dk} (\texttt{prompt-1}, \texttt{prompt-3}, \texttt{prompt-5}, \texttt{prompt-7}, \texttt{prompt-9}) or Machine Learning Feature Importance (MLFI) refers to the scenario where important features are identified by ML algorithms such as \texttt{XGB}, \texttt{RF}, \texttt{Ada}, \texttt{LR}, \texttt{Ensemble}, and `Even dk' (\texttt{prompt-2}, \texttt{prompt-4}, \texttt{prompt-6}, \texttt{prompt-8}, \texttt{prompt-10}) or MLFI-ord considers the order of features in addition to their importance.

\begin{table}[!t]
\caption{Summary of Domain Knowledge Types}
\centering
\begin{tabularx}{\textwidth}{>{\raggedright\arraybackslash}p{1.5cm} >{\raggedright\arraybackslash}p{2cm} >{\raggedright\arraybackslash}p{3.5cm} >{\raggedright\arraybackslash}p{2cm} >{\raggedright\arraybackslash}p{3cm}}
\toprule
\textbf{DK} & \textbf{Name} & \textbf{Description} & \textbf{Focus} & \textbf{Implementation} \\
\midrule
dk0 & N/A & No extra domain knowledge & N/A & Solely data-driven \\
\addlinespace
Odd dk & \small{MLFI} & ML defines feature importance & Feature Attribution & \texttt{XGB, RF, Ada, LR, Ensemble}\\
\addlinespace
Even dk & \small{MLFI-ord} & Similar to MLFI, includes feature order & Feature Attribution & \texttt{XGB, RF, Ada, LR, Ensemble} \\
\bottomrule
\end{tabularx}
\label{tab:my_label}
\end{table}

\section{Experimental Setup}

\textbf{Task.} This work focuses on binary classification within a credit assessment context in machine learning (ML). The task involves learning a function $f: \mathcal{X} \rightarrow {0,1}$, predicting a binary outcome $y \in {0,1}$ for each feature instance $x \in \mathcal{X}$. The feature space, $\mathcal{X}$, comprises a protected attribute $G$ (e.g., age, race, sex) and all other attributes, $\mathcal{X'}$. Together, they form the feature vector for each instance, $\mathcal{X} = (G, \mathcal{X'})$. The outcome, $y$, denotes an individual's creditworthiness.

\vspace{1.5mm}
\noindent \textbf{Hyperparameters and Models.} We employed six ML models, each with a distinct set of hyperparameters. These were optimized using a randomized search cross-validation (CV) strategy, experimenting with 25 unique hyperparameters. This led to an extensive model tuning process involving numerous model iterations. We used a 5-fold CV (0.8, 0.2), with \textit{RandomizedSearchCV} over 20 iterations. The exact hyperparameters depended on the specific model:

\begin{itemize}
  \item \RandomForest: `n-estimators', `max-depth', `min-samples-split', `min-samples-leaf', `bootstrap'. (Total = 5)
  \item \LogisticRegression: `C', `penalty', `solver'. (Total = 3)
  \item \MLPClassifier: `hidden-layer-sizes', `activation', `solver', `alpha', `learning-rate', `max-iter'. (Total = 6)
  \item \KNeighbors: `n-neighbors', `weights', `algorithm', `leaf-size', `p'. (Total = 5)
  \item \XGBoost: `n-estimators', `l-rate', `max-depth', `colsample-bytree'. (Total = 4)
  \item \AdaBoost: `n-estimators', `learning-rate'. (Total = 2)
\end{itemize}

\noindent \textbf{Dataset.} We used the German Credit dataset, a space-efficient choice with 1,000 individuals and 21 attributes for creditworthiness classification. Cleaned by Le et al.\cite{le2022survey} ), this dataset aids banks in lending decisions, using gender as a fairness-sensitive feature.

\noindent \textbf{Bootstrap sampling.} To address imbalances and distribution disparities between groups (e.g., Male vs. Female), we employed bootstrapping with 1000 resamples. Bootstrapping is a robust statistical technique that estimates sampling distributions through data resampling. By generating resampled datasets, calculating the mean disparity (here TPR) for each, and analyzing the resulting distributions, we assessed the statistical significance of the observed difference.
\vspace{-5mm}
\section{Results}

\begin{table}[ht]
\centering
\caption{Performance Comparison of Models on the German Credit Dataset. Average accuracy results in the classical-ML part are computed excluding random.}
\begin{tabularx}{0.8\textwidth}{lllcccccc}
\toprule
Model & \makecell{DK \\ Type} &\makecell{ML \\ model}& Pre.$\uparrow$ & Rec.$\uparrow$ & F1$\uparrow$ & \makecell{FP\\ Cost}$\downarrow$ & \makecell{FN \\ Cost}$\downarrow$ \\
\midrule
\textbf{\texttt{RF}} & - & - & \textbf{\underline{0.8153}} & 0.9078 & \textbf{\underline{0.8591}} & 145.0 & 13.0 \\
\texttt{LR} & - & - & 0.7368 & 0.8936 & 0.8077 & 225.0 & 15.0 \\
\texttt{MLP} & - & - & 0.7654 & 0.8794 & 0.8185 & 190.0 & 17.0 \\
\texttt{KNN} & - & - & 0.7707 & 0.8582 & 0.8121 & 180.0 & 20.0 \\
\texttt{XGB} & - & - & 0.8077 & \underline{0.8936} & 0.8485 & 150.0 & 15.0 \\
\texttt{AdaBoost} & - & - & 0.7875 & 0.8936 & 0.8372 & 170.0 & 15.0 \\
\texttt{random} & - & - & 0.7625 & 0.4326 & 0.5520 & 95.0 & 80.0 \\
\midrule
\textit{Avg.} & - & - & 0.7792 & 0.8822 & 0.8302 & 172.5 & 15.6 \\
\midrule
\texttt{prompt-0} & \scriptsize{N/A} & \texttt{-} & 0.7625 & 0.4326 & 0.5520 & 95.0 & 80.0 \\
\texttt{prompt-1} & \scriptsize{MLFI} & \texttt{XGB} & 0.7083 & 0.7234 & 0.7158 & 210.0 & 39.0 \\
\texttt{prompt-2} & \scriptsize{MLFI-ord} & \texttt{XGB} & 0.6842 & 0.5532 & 0.6118 & 180.0 & 63.0 \\
\texttt{prompt-3} & \scriptsize{MLFI}  & \texttt{RF} & 0.7206 & 0.6950 & 0.7076 & 190.0 & 43.0 \\
\texttt{prompt-4} & \scriptsize{MLFI-ord} & \texttt{RF} & 0.7087 & 0.5177 & 0.5984 & 150.0 & 68.0 \\
\textbf{\color{blue}{\texttt{prompt-5}}} & \scriptsize{MLFI}  & \texttt{Ada} & \textbf{\color{blue}{0.7305}} & \textbf{\color{blue}{0.7305}} & \textbf{\color{blue}{0.7305}} & 190.0 & 38.0 \\
\texttt{prompt-6} & \scriptsize{MLFI-ord} & \texttt{Ada} & 0.7404 & 0.5461 & 0.6286 & 135.0 & 64.0 \\
\texttt{prompt-7} & \scriptsize{MLFI}  & \texttt{LR} & 0.7154 & 0.6596 & 0.6863 & 185.0 & 48.0 \\
\texttt{prompt-8} & \scriptsize{MLFI-ord} & \texttt{LR} & 0.6957 & 0.4539 & 0.5494 & 140.0 & 77.0 \\
\texttt{prompt-9} & \scriptsize{MLFI}  & \texttt{ensemble} & 0.7209 & 0.6596 & 0.6889 & 180.0 & 48.0 \\
\texttt{prompt-10} & \scriptsize{MLFI-ord} & \texttt{ensemble} & 0.7037 & 0.5390 & 0.6104 & 160.0 & 65.0 \\
\midrule
\textit{Avg.} & - & - & 0.7129 & 0.6078 & 0.6528 & 176.5 & 56.3 \\
\midrule
\bottomrule
\end{tabularx}
\label{tab:my_label}
\end{table}

\textbf{Accuracy.} Table \ref{tab:my_label} presents a comparative analysis of the performance of various models equipped with different types of domain knowledge and machine learning algorithms. The performance metrics under consideration include precision (Pre.), recall (Rec.), F1 score (F1), false-positive cost (FP Cost), and false-negative cost (FN Cost). Given the context of credit risk assessment, the latter two metrics bear particular importance, with the false-positive cost being assigned a weight of 5 to reflect the higher financial risk associated with erroneously granting credit to an unworthy applicant.

OpenAI-based models are tested under different scenarios using Machine Learning Feature Importance (MLFI) and its ordered variant (MLFI-ord), which can be seen as an attempt to incorporate domain knowledge into the model. Notably, \texttt{$\text{Prompt}-{5}$}, using \AdaBoost model under the MLFI scenario, achieves the highest precision, recall, and F1 score among all OpenAI-based models, which are 0.7305 in each metric. This suggests that using a combination of domain knowledge (MLFI) and the \AdaBoost model, the OpenAI-based model can achieve balanced and competitive results (the results are comparable with LR, XGB, and \AdaBoost in terms of Pre). Overall, when utilizing \AdaBoost and \RandomForest 
 as domain knowledge, we observe relatively high performance. It is noteworthy that these models performed exceptionally well in the classical ML part, especially considering their F1 scores. To our surprise, we did not observe a significant advantage when using an ordered feature introduction. Instructing ChatGPT to use ordered feature values often led to poorer performance in many cases.

However, when we compare the average values of classical models with those of OpenAI-based models, we see that the former outperforms the latter in all accuracy metrics: precision (0.7792 vs. 0.7129), recall (0.8822 vs. 0.6078), and F1 score (0.8302 vs. 0.6528). This suggests that, for this specific task, classical models are generally more effective. Importantly, it's worth noting that the classical machine learning models used approximately 80\% of the available data (i.e., \squotes{800} samples) for training, while the OpenAI-ML models only utilized 20 samples as training examples. This means that the classical models had a significant data advantage, using the information at a rate \underline{40 times greater} than that of OpenAI-ML. Despite this disparity in data, the OpenAI-ML still produced competitive results, highlighting their potential and efficiency. Additionally, it is worth considering that OpenAI-ML offers the unique advantage of producing human-controllable outputs in the form of prompts that can be generated through a predefined course of action. An intriguing observation is the reduced False Positive (FP) cost associated with OpenAI models compared to traditional models—an aspect of utmost importance in credit risk assessment. This suggests that OpenAI models demonstrate a cautious approach in averting certain false alarms, a trait that could potentially be amplified through targeted instruction. However, this cautious stance might also make them more susceptible to overlooking true instances.

\begin{table}[!h]
    \centering
    \caption{Gender fairness results based on true-positive-rate. Note that the TPR values for each group, obtained via bootstrap sampling, are represented by $\mathbb{E}[\mathcal{E}]$. The disparity in the true positive rate is denoted by $\Delta$.}
    \begin{tabular}{lcccc}
    \toprule
    \texttt{clf} & \multicolumn{2}{c}{Sex} & \multicolumn{2}{c}{$\Delta$ and $H_0$} \\
    \cmidrule(lr){2-3} \cmidrule(lr){4-5}
    & $\mathbb{E}[\mathcal{E}]_{M}$ & $\mathbb{E}[\mathcal{E}]_{F}$ & $\Delta_{g}$ & $\text{reject} \ H_0$ \\
    \midrule
    \RandomForest & 0.6611 & 0.6325 & 0.0286 & \textcolor{red}{\textbf{True}} \\
    \LogisticRegression & 0.6419 & 0.6240 & 0.0178 & \textcolor{red}{\textbf{True}} \\
    \MLPClassifier & 0.6254 & 0.6181 & 0.0073 & \textcolor{red}{\textbf{True}} \\
    \KNeighbors & 0.5709 & 0.6190 & -0.0486 & \textcolor{red}{\textbf{True}} \\
    \XGBoost & 0.6382 & 0.6267 & 0.0116 & \textcolor{red}{\textbf{True}} \\
    \AdaBoost & 0.6057 & 0.6401 & -0.0344 & \textcolor{red}{\textbf{True}} \\
    \texttt{Prompt-1} & 0.5156 & 0.5070 & 0.0085 & \textcolor{red}{\textbf{True}} \\
    \texttt{Prompt-3} & 0.5709 & 0.4587 & 0.1122 & \textcolor{red}{\textbf{True}} \\
    \texttt{Prompt-5} & 0.5200 & 0.5143 & 0.0063 & False \\
    \texttt{Prompt-7} & 0.4670 & 0.4641 & 0.0031 & False \\
    \texttt{Prompt-9} & 0.4485 & 0.4706 & -0.0221 & \textcolor{red}{\textbf{True}} \\
    \bottomrule
    \end{tabular}
\end{table}
\noindent \textbf{Fairness.} The fairness analysis of the provided results, which leverages \dquotes{odd dk} due to its superior preceding task performance, offers insightful conclusions. In contrast to machine learning models, certain prompts could achieve fair outcomes, i.e., a non-significant difference in the efforts required by different genders. Despite comparable performance to prompts, machine learning models, including \RandomForest, \LogisticRegression, \MLPClassifier, \KNeighbors, \XGBoost, and \AdaBoost, consistently rejected the null hypothesis, implying significant gender-based effort disparity and lack of fairness. Prompts, however, showcased more diverse results. For instance, \texttt{Prompt-5} and \texttt{Prompt-7} suggested relative fairness, whereas \texttt{Prompt-1}, \texttt{Prompt-3}, and \texttt{Prompt-9} indicated significant effort differences. Interestingly, some prompts like  \texttt{Prompt-9} even reversed disparity direction, favoring the less privileged group. This underlines the potential of prompts to promote fairness, even if they do not always yield statistically significant results. Our analysis emphasizes the need for a holistic model evaluation approach, going beyond statistical metrics, to encompass performance and fairness implications for different demographics.
\vspace{-2mm}
\section{Conclusion} The study exhibits the benefits of Large-scale Language Models (LLMs), in particular OpenAI's ChatGPT, in Machine Learning tasks. It uses prompt engineering to optimize behavior and prediction accuracy, suggesting that these models could potentially perform comparably, if not better, than traditional ML models. Integration of domain knowledge shows an interesting impact on accuracy, and gender fairness enhancement, setting the basis for broader future investigations. Future exploration requires prompt design optimization, providing ample reasoning time for the system systems through concepts such as Chain-of-Thought Prompting \cite{wei2022chain}, and fine-tuning methodologies. The incorporation of methods for system decision explanation, as well as the application of GPT-based systems to recommender systems while mitigating biases, are critical considerations for further exploration \cite{zhang2023chatgpt,wang2023decodingtrust,naghiaei2022cpfair,amigo2023unifying,deldjoo2023fairness}.

\bibliographystyle{splncs04}
\bibliography{mybibliography}

\begin{thebibliography}{10}
\providecommand{\url}[1]{\texttt{#1}}
\providecommand{\urlprefix}{URL }
\providecommand{\doi}[1]{https://doi.org/#1}

\bibitem{alnuhait2023facechat}
Alnuhait, D., Wu, Q., Yu, Z.: Facechat: An emotion-aware face-to-face dialogue
  framework. arXiv preprint arXiv:2303.07316  (2023)

\bibitem{amigo2023unifying}
Amig{\'o}, E., Deldjoo, Y., Mizzaro, S., Bellog{\'\i}n, A.: A unifying and
  general account of fairness measurement in recommender systems. Information
  Processing \& Management  \textbf{60}(1),  103115 (2023)

\bibitem{brown2020language}
Brown, T., Mann, B., Ryder, N., Subbiah, M., Kaplan, J.D., Dhariwal, P.,
  Neelakantan, A., Shyam, P., Sastry, G., Askell, A., et~al.: Language models
  are few-shot learners. Advances in neural information processing systems
  \textbf{33},  1877--1901 (2020)

\bibitem{chang2023survey}
Chang, Y., Wang, X., Wang, J., Wu, Y., Zhu, K., Chen, H., Yang, L., Yi, X.,
  Wang, C., Wang, Y., et~al.: A survey on evaluation of large language models.
  arXiv preprint arXiv:2307.03109  (2023)

\bibitem{chowdhery2022palm}
Chowdhery, A., Narang, S., Devlin, J., Bosma, M., Mishra, G., Roberts, A.,
  Barham, P., Chung, H.W., Sutton, C., Gehrmann, S., et~al.: Palm: Scaling
  language modeling with pathways. arXiv preprint arXiv:2204.02311  (2022)

\bibitem{clavie2023large}
Clavi{\'e}, B., Ciceu, A., Naylor, F., Souli{\'e}, G., Brightwell, T.: Large
  language models in the workplace: A case study on prompt engineering for job
  type classification. In: International Conference on Applications of Natural
  Language to Information Systems. pp. 3--17. Springer (2023)

\bibitem{deldjoo2023fairness}
Deldjoo, Y., Jannach, D., Bellogin, A., Difonzo, A., Zanzonelli, D.: Fairness
  in recommender systems: research landscape and future directions. User
  Modeling and User-Adapted Interaction pp. 1--50 (2023)

\bibitem{le2022survey}
Le~Quy, T., Roy, A., Iosifidis, V., Zhang, W., Ntoutsi, E.: A survey on
  datasets for fairness-aware machine learning. Wiley Interdisciplinary
  Reviews: Data Mining and Knowledge Discovery  \textbf{12}(3),  e1452 (2022)

\bibitem{li2023fairness}
Li, Y., Zhang, Y.: Fairness of chatgpt. arXiv preprint arXiv:2305.18569  (2023)

\bibitem{naghiaei2022cpfair}
Naghiaei, M., Rahmani, H.A., Deldjoo, Y.: Cpfair: Personalized consumer and
  producer fairness re-ranking for recommender systems. In: Proceedings of the
  45th International ACM SIGIR Conference on Research and Development in
  Information Retrieval. pp. 770--779 (2022)

\bibitem{qiu2023smile}
Qiu, H., He, H., Zhang, S., Li, A., Lan, Z.: Smile: Single-turn to multi-turn
  inclusive language expansion via chatgpt for mental health support. arXiv
  preprint arXiv:2305.00450  (2023)

\bibitem{schaeffer2023emergent}
Schaeffer, R., Miranda, B., Koyejo, S.: Are emergent abilities of large
  language models a mirage? arXiv preprint arXiv:2304.15004  (2023)

\bibitem{DBLP:journals/corr/abs-2201-08239}
Thoppilan, R., Freitas, D.D., Hall, J., Shazeer, N., Kulshreshtha, A., Cheng,
  H., Jin, A., Bos, T., Baker, L., Du, Y., Li, Y., Lee, H., Zheng, H.S.,
  Ghafouri, A., Menegali, M., Huang, Y., Krikun, M., Lepikhin, D., Qin, J.,
  Chen, D., Xu, Y., Chen, Z., Roberts, A., Bosma, M., Zhou, Y., Chang, C.,
  Krivokon, I., Rusch, W., Pickett, M., Meier{-}Hellstern, K.S., Morris, M.R.,
  Doshi, T., Santos, R.D., Duke, T., Soraker, J., Zevenbergen, B., Prabhakaran,
  V., Diaz, M., Hutchinson, B., Olson, K., Molina, A., Hoffman{-}John, E., Lee,
  J., Aroyo, L., Rajakumar, R., Butryna, A., Lamm, M., Kuzmina, V., Fenton, J.,
  Cohen, A., Bernstein, R., Kurzweil, R., y~Arcas, B.A., Cui, C., Croak, M.,
  Chi, E.H., Le, Q.: Lamda: Language models for dialog applications. CoRR
  \textbf{abs/2201.08239} (2022), \url{https://arxiv.org/abs/2201.08239}

\bibitem{wang2023decodingtrust}
Wang, B., Chen, W., Pei, H., Xie, C., Kang, M., Zhang, C., Xu, C., Xiong, Z.,
  Dutta, R., Schaeffer, R., et~al.: Decodingtrust: A comprehensive assessment
  of trustworthiness in gpt models. arXiv preprint arXiv:2306.11698  (2023)

\bibitem{wei2022chain}
Wei, J., Wang, X., Schuurmans, D., Bosma, M., Xia, F., Chi, E., Le, Q.V., Zhou,
  D., et~al.: Chain-of-thought prompting elicits reasoning in large language
  models. Advances in Neural Information Processing Systems  \textbf{35},
  24824--24837 (2022)

\bibitem{zhang2023chatgpt}
Zhang, J., Bao, K., Zhang, Y., Wang, W., Feng, F., He, X.: Is chatgpt fair for
  recommendation? evaluating fairness in large language model recommendation.
  arXiv preprint arXiv:2305.07609  (2023)

\end{thebibliography}

\end{document}